\documentclass{egpubl}
\usepackage{eg2013}

 \ConferencePaper        %
 \electronicVersion %

\ifpdf \usepackage[pdftex]{graphicx} \pdfcompresslevel=9
\else \usepackage[dvips]{graphicx} \fi

\PrintedOrElectronic

\usepackage{t1enc,dfadobe}

\usepackage{egweblnk}
\usepackage{cite}

\usepackage{subfig}

\usepackage{float}

\usepackage{todonotes}
\reversemarginpar

\usepackage{booktabs}

\usepackage{pgfplots}
\usetikzlibrary{spy, shapes, arrows, matrix, pgfplots.groupplots}

\pdfpageattr{/Group <</S /Transparency /I true /CS /DeviceRGB>>} 

\usepackage{amsmath}
\usepackage{amssymb}

\DeclareMathOperator*{\argmin}{arg\,min}

\usepackage{adjustbox}
\newcommand\mygraphics[1] {}
\newcommand\mygraphicsb[1] {}

\newcommand{\mycomment}[1] {}

\renewcommand{\paragraph}[1] {\noindent \textbf{#1} \ }

\newcommand{\chapt}[1] {
\section{#1}
}

\newcommand{\sect}[1] {
\subsection{#1}
}

\newcommand{\subsect}[1] {
\subsubsection{#1}
}

 \newcommand{\subsubsect}[1] {
 \paragraph{#1}
 }

\title[DuctTake]{DuctTake: Spatiotemporal Video Compositing}
\vspace{-.2cm}
\ifdefined\IsSubmission
\author[1101]
       {Submission ID: 1101}
\else
\author[Jan R\"uegg, Oliver Wang, Aljoscha Smolic, Markus Gross]
       {Jan R\"uegg$^{1,2}$, Oliver Wang$^{2}$, Aljoscha Smolic$^2$, Markus Gross$^{1,2}$
        \\
         $^1$ETH Zurich, Switzerland\\
         $^2$Disney Research Zurich, Switzerland\\         
       }
\fi

\begin{document}

\teaser{
\vspace{-.2cm}
\centering
\newcommand\graphic[1]{\includegraphics[height=2.2cm]{#1}}
\vspace{-1cm}
\subfloat[][]{\graphic{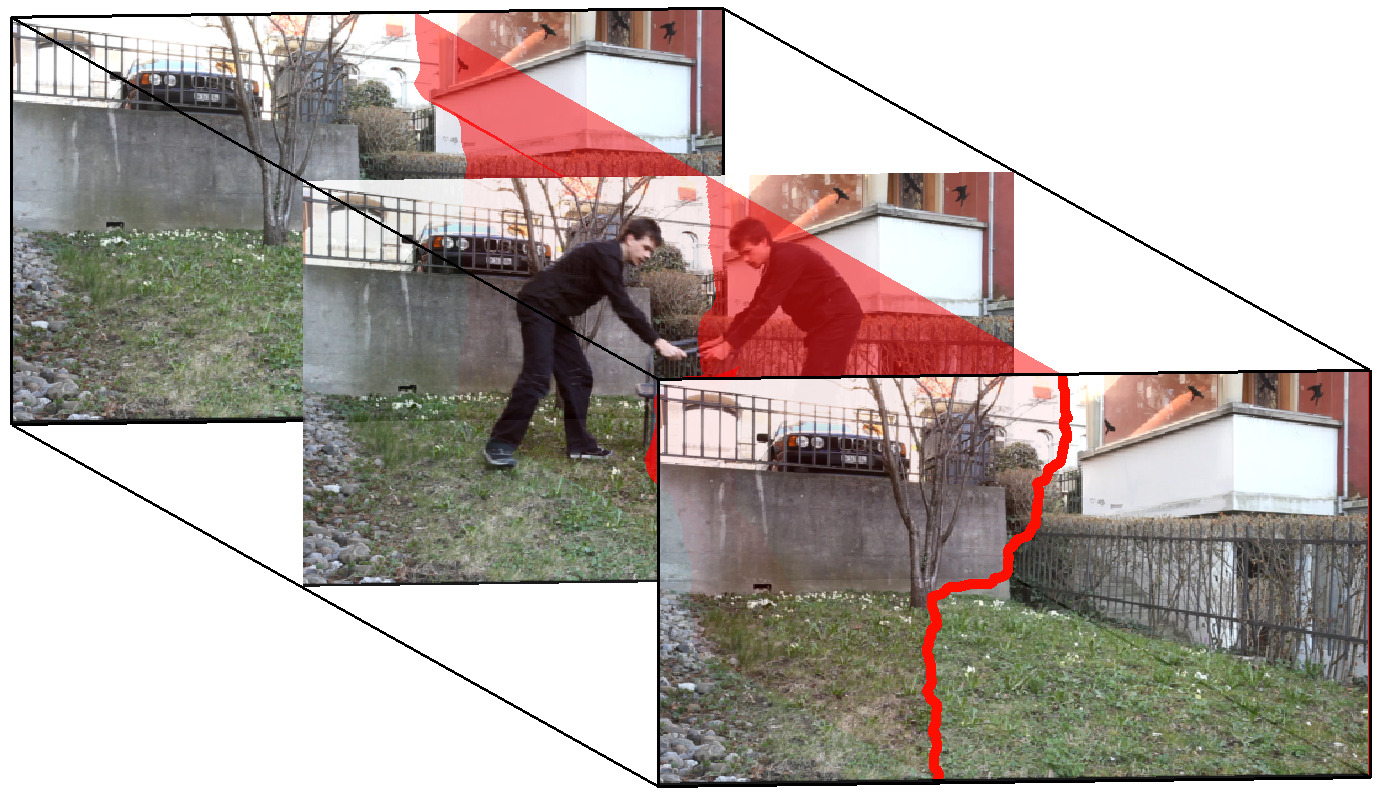}} \
\subfloat[][]{\graphic{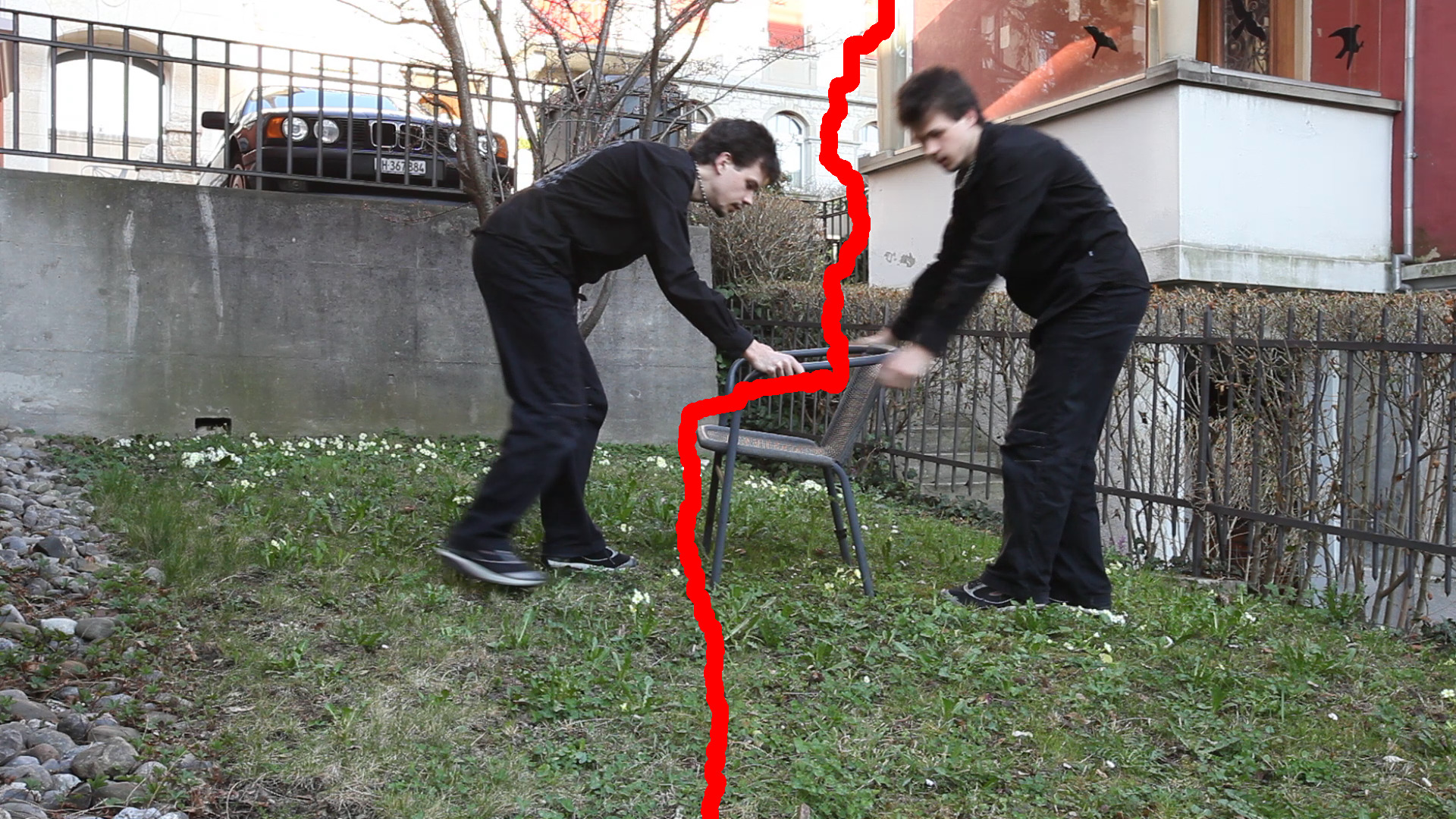}} \ 
\subfloat[][]{\graphic{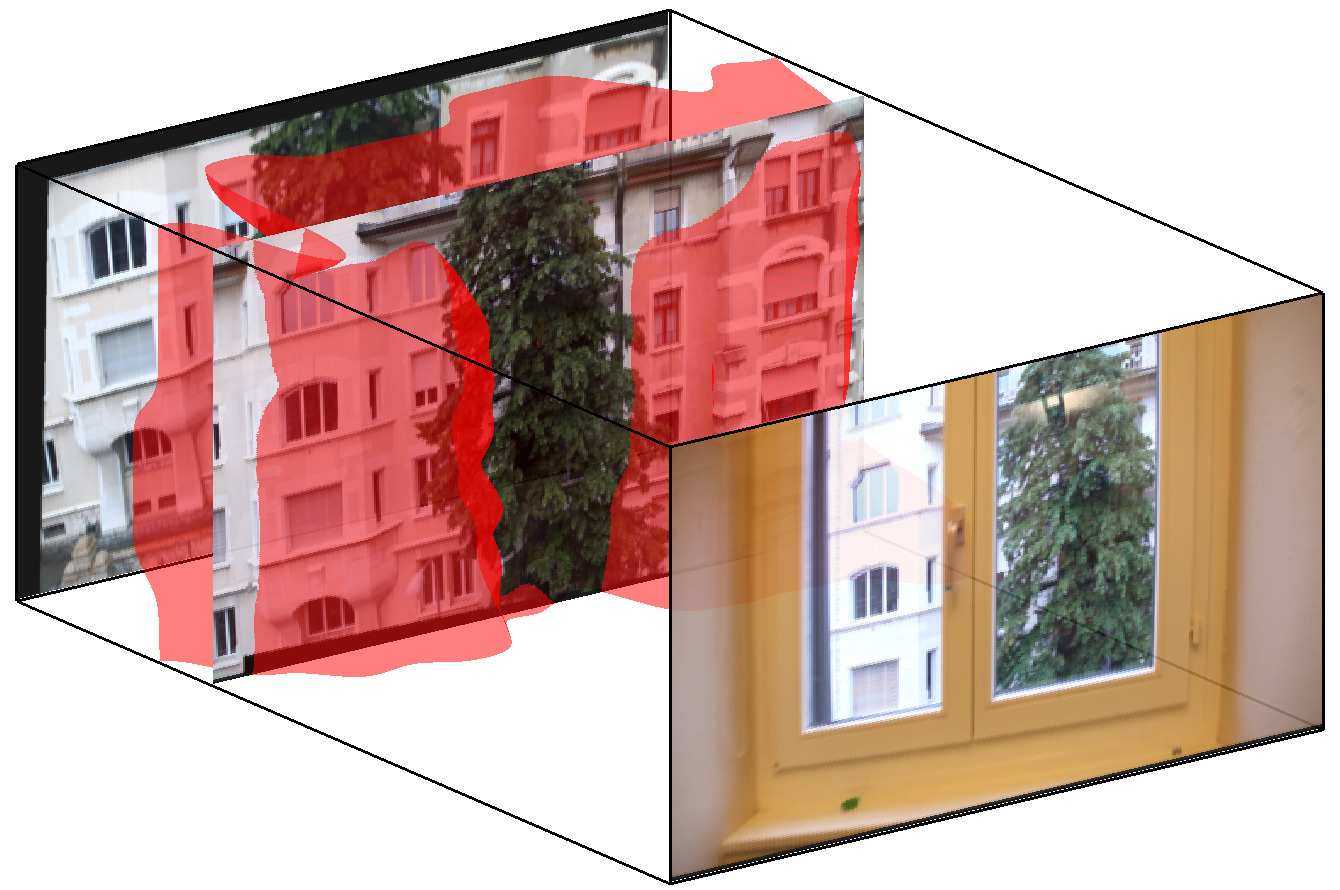}} \
\subfloat[][]{\graphic{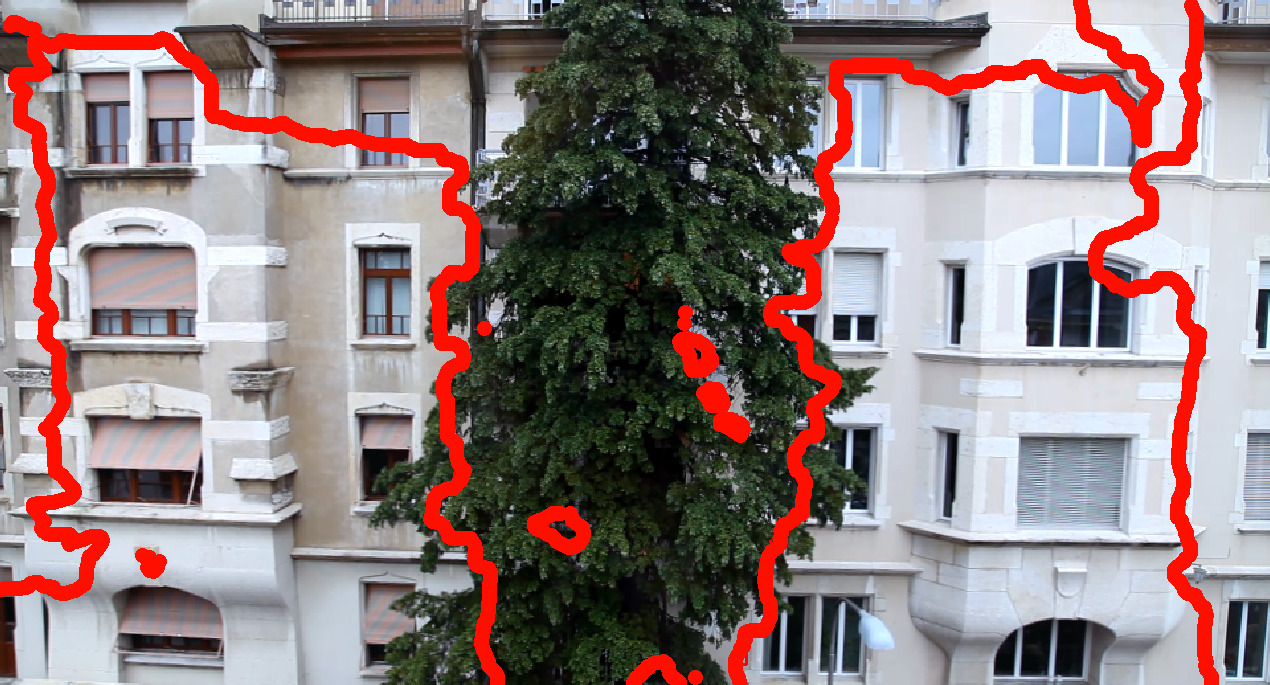}} \
\caption{Two examples of spatio-temporal seams (red) computed to optimally composite a pair of videos (a,c). Each seam projected onto a single frame shows complex, non-intuitive shapes that achieve high quality, stable composites (b,d).}
\label{fig:teaser}

}

\maketitle

\begin{abstract}
DuctTake is a system designed to enable practical compositing of multiple takes of a scene into a single video. 
Current industry solutions are based around object segmentation, a hard problem that requires extensive manual input and cleanup, making compositing an expensive part of the film-making process.  
Our method instead composites shots together by finding optimal spatiotemporal seams using motion-compensated 3D graph cuts through the video volume.
We describe in detail the required components, decisions, and new techniques that together make a usable, interactive tool for compositing HD video, paying special attention to running time and performance of each section. 
We validate our approach by presenting a wide variety of examples and by comparing result quality and creation time to composites made by professional artists using current state-of-the-art tools.
   

\end{abstract}

\chapt{Introduction}

Assembling images composed of multiple photographs is as old as photography itself.
Originally achieved through arduous manual cut-and-paste, advanced digital tools now exist that make photo compositing easier (e.g., Poisson blending, alpha matting, graph cuts, etc).
However, video compositing remains a challenging problem, as additional difficulties such as increased computational requirements, temporal stability, and alignment make video extensions of photographic methods non-trivial. 
Nonetheless, compositing video is an integral part of modern film making, and virtually all big-budget movies contain a number of scenes composed of multiple sources.
Uses include special effects shots, realistic background replacement, combining optimal actor performances from multiple takes and removing unwanted scene elements or mistakes.
Video compositing is still most commonly accomplished by the digital equivalent of ``cut-and-paste'', rotoscoping, or by chroma-keying. 
While chroma keying is robust and cheap, it cannot be used in all cases as it greatly restricts filming environments and often requires challenging color balancing in post production. 
On the other hand, rotoscoping is largely manual, expensive and time consuming and therefore is most commonly used only for expensive effect shots.

We address the problem of video compositing using a different approach.
Instead of segmenting objects, we present DuctTake (named for duct-taping together two takes), a collection of algorithms and workflows for finding optimal space-time ``seams'' to join together videos. 
Because this approach solves a simpler problem, it is robust, fast to compute, and easy for artists to use, enabling compositing techniques to be used on lower budget shots and productions.
This simplification of problem domain comes at the expense of stricter prerequisites for the kinds of video clips that can be composited.
Instead of compositing content from arbitrary clips, we instead focus on combining multiple \emph{takes} together, where the content and camera location are expected to be similar; importantly, there must be enough shared scene content for low-visibility seams to exist. 

The main contribution of our work is the presentation and description of an intuitive, efficient \emph{system} and \emph{workflow} for compositing together two FullHD video sources.
This task involves difficult steps of video alignment, content matching, and motion-compensation.
While individual components of this system address known problems in video processing, we believe that a detailed description of our working system will provide insight into which of the many available solutions can be practically used, especially considering strict requirements for memory usage and running time.
In addition, we present several non-intuitive technical contributions, where our fast and simple methods outperformed those commonly used in the literature, including: 
\begin{itemize}
\item{Computation of spatiotemporal seams by coarse-to-fine graph cuts in a \emph{motion-compensated} videocube.}
\item{A robust and fast alignment technique that combines hierarchical block-based matching with drift-corrected alignment propagation.}
\item{Efficient approximations to matching blur kernel across takes and finding optimal cropping volumes.}
\end{itemize}
We validated our approach by creating a wide range of examples (available in the supplementary material along with screen-casts of their creation), and by directly comparing the quality of our results and time of creation to composites produced by professional artists using state-of-the-art tools. 
\chapt{Related Work}
\newcommand\related[1]{\noindent \textbf{#1} \ }
Our work encompasses many research areas, and builds on a large amount of prior work. 
Specific techniques that we used will be referenced in the appropriate implementation sections. 
Here we will discuss other works with goals similar to ours. 

\related{Object Segmentation}
Compositing is often accomplished by segmenting objects from images or videos.
Numerous techniques exist where user input is leveraged to help resolve object boundaries (e.g., brush strokes~\cite{wang2005interactive,tong2011video}, rotoscoping~\cite{agarwala2004keyframe}, or alpha matting via trimaps~\cite{chuang2002video,wang2007soft}).
Once isolated, objects can then be pasted into desired scenes.
While these methods do not require common visual elements in both videos, computing precise object boundaries is a very difficult problem.
Our method instead finds seams to join similar videos.
This approach has stricter content requirements, but is more robust to difficult segmentation problems such as blurred object boundaries and texture ambiguities.

\related{Seams}
Probably most similar to our work is \emph{Interactive digital photomontage}\cite{agarwala2004interactive}, which combines multiple photographs into a single output with a few simple strokes. 
We extend this idea to video sequences, which introduces a host of new problems such as computational tractability, temporal consistency, and alignment. 
Video compositing has also been addressed by computing seams in the gradient domain~\cite{wang2007videoshop}.
Gradient compositing can yield good results, but requires solving a computationally expensive integration of the gradient video volume.
In addition, gradient compositing methods are highly seam dependent, and work best only with hand-chosen seams that run through largely untextured regions, as differences on the seams cause color bleeding.

\related{Video cuts}
Graph cuts through video volumes have been used in the past for applications with different goals, such as automatic video-texture generation, or content extraction.
\emph{Graphcut Textures}~\cite{kwatra2003graphcut} computes video textures by finding optimal seams to loop video, but did not give users the possibility to selectively choose objects from different videos. 
\emph{Space-Time Video Montage}~\cite{kang2006space} performs automatic video summarization by graph cuts, and as such does not share the same goal of user-driven seamless compositing.
\emph{Panorama video textures}~\cite{agarwala2005panoramic}
and \emph{Selectively De-animating Video}~\cite{bai2012selectively} both produce composited videos, however, they generate results with mostly static content or small looped motions.
DuctTake on the other hand is a general-purpose compositing tool driven by an interactive interface. 
Most significantly, it is designed to operate where foreground and background can be moving arbitrarily, which introduces difficult alignment, motion compensation, and temporal stability issues. 

\related{Alignment}
A key component of our system is a new block-based propagate-and-refine matching technique, that is very fast, and yields temporally stable and robust results.
Commonly, existing approaches align videos by feature matching~\cite{sand2004video}.
We instead use a hierarchical compass-search, which leverages similar local search strategies widely used in the video coding community~\cite{tham1998novel,zhu2000new}.
These approaches have a fundamentally different goal; they are designed for frame-to-frame matching and need only to minimize block differences and reduce entropy~\cite{zhu2000new}.
As a result, they have problems with aligning different views with large differences in content~\cite{tham1998novel}.
Our approach instead imposes a coarse-to-fine scheme which instead of merging blocks with similar motion~\cite{kim1994hierarchical}, matches large image regions deep in the pyramid, creating smoother, spatially consistent results. 
This combined with a propagate-and-refine approach yields temporally consistent alignments while still allowing for large differences in camera positions and content. 
\chapt{Algorithmic Framework}

\tikzstyle{block} = [rectangle, draw, fill=blue!20,
    text width=5.5em, text centered, rounded corners, minimum height=2em]
\tikzstyle{blockreq} = [rectangle, draw, fill=red!20,
    text width=5.5em, text centered, rounded corners, minimum height=2em]
\tikzstyle{line} = [draw, -latex']
\tikzstyle{cloud} = [draw, ellipse,fill=red!20, minimum height=2em]

\begin{figure}[h]
\centering
\begin{tikzpicture}[]
\matrix [column sep=0em,row sep=0.7em] {
    \node [cloud] (video1) {Video $A$}; & & \node [cloud] (video2) {Video $B$}; \\
    & & \node [block] (align2) {Align (\ref{alignment})}; \\
    \node [block] (blur1) {Blur (\ref{blur})}; & & \node [block] (blur2) {Blur (\ref{blur})}; \\
    & \node [blockreq] (cut) {Cut (\ref{cut})}; & \node [block] (color2) {Color (\ref{color})}; \\
    & \node [block] (blend) {Blend (\ref{blend})}; & \\
    & \node [block] (crop) {Crop (\ref{crop})}; & \\
    & \node [cloud] (final) {Output Video}; & \\
};

    \path [line] (video1) |- (align2);
    \path [line] (video2) -- (align2);
    \path [line] (video1) -- (blur1);
    \path [line] (align2) -- (blur2);
    \path [line] (blur2) -- (color2);
    \path [line] (blur1) |- (cut);
    \path [line] (blur1) |- (blend);
    \path [line] (color2) |- (blend);
    \path [line] (color2) -- (cut);
    \path [line] (cut) -- (blend);
    \path [line] (blend) -- (crop);
    \path [line] (crop) -- (final);

\end{tikzpicture}
\caption{\label{fig:overview}Algorithm overview, blue blocks are optional.}
\end{figure}
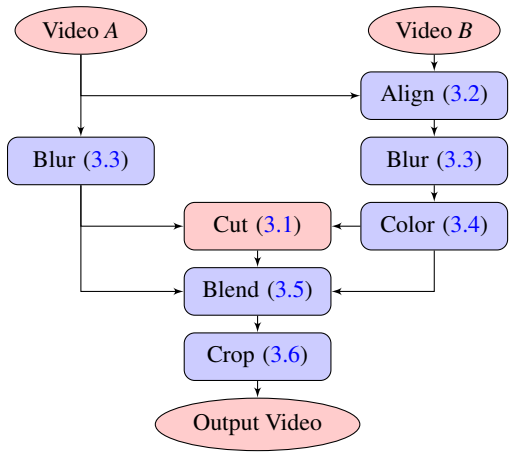

An overview of our method is presented in Figure~\ref{fig:overview} with corresponding sections in the text.
The user provides a few high-level sources of information; the reference video to match colors with, the temporal offset required to synchronize events that should occur at the same time, and a few quick brush strokes indicating the parts of the video to keep from each take.
The algorithm then computes an optimal seam and merges the two videos together.
All performance measurements that we give throughout the paper were made with a 64bit Intel Core i7 CPU system on 1080p videos unless otherwise specified. 
Videos referenced in the paper are written in \textsc{SmallCaps} and can be viewed in the supplemental material under the same name.

The core technique for computing these seams is a straightforward video extension to the classic image labeling problem, for which graph cuts have long been used as a solution~\cite{boykov2001fast}. 
However, several additions are required to handle difficulties of working with video, specifically alignment due to camera motion, temporal stability, and efficient content matching, discussed in the following sections. 
For our application, each pixel is assigned a label that indicates which source videos it is from ($A$ or $B$).
A ``seam'' exists at locations where labels change between neighbors. 
Figure~\ref{fig:teaser} shows two of these seams (red) that are computed using a 3D spatiotemporal graph cut (sequences \textsc{Chair} and \textsc{ThroughWindow}).
While many of the examples that we will present involve spatial cuts or temporal blends, seams can form any complex, disconnected space-time manifold.

\sect{Video Graph Cuts}
\label{cut}

\begin{figure}[ht]
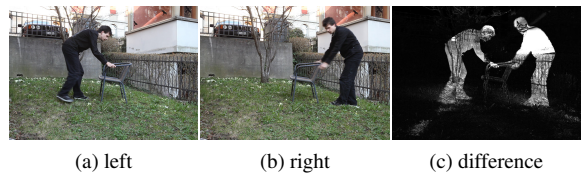

  \newcommand\graphics[1]{\adjincludegraphics[width=.33\linewidth,trim={{.1\width} {0\height} {.1\width} {0\height}},clip]{#1}}
  \centering
  \subfloat[][left]{\label{fig:chair1}\graphics{./figures/chair1}}\hfill
  \subfloat[][right]{\label{fig:chair2}\graphics{./figures/chair2}}\hfill
  \subfloat[][difference]{\label{fig:chair3}\graphics{./figures/chair3}}
  \caption{Difference map for two frames.}
  \label{fig:chairs}
\end{figure}

Consider Figure~\ref{fig:chairs} showing the \textsc{Chair} dataset.
The goal is to composite the same person from two takes.
To complicate things, the person exchanges a common object (the chair).
To avoid duplication of the chair, the desired seam must pass through the object during the video. 
The difference map (Figure~\ref{fig:chair3}), intuitively shows us similar areas, where seams will be least visible (darker regions). 
We introduce a seam penalty equal to the sum of the squared distance between the colors of the first and second video ($A,B$) at each pixel it separates.
Our goal is to find a seam that separates strokes with the minimum visibility penalty.

\newcommand\graphics[1]{\adjincludegraphics[width=.33\linewidth,trim={{.1\width} {0\height} {.1\width} {0\height}},clip]{#1}}

\begin{figure}[t]
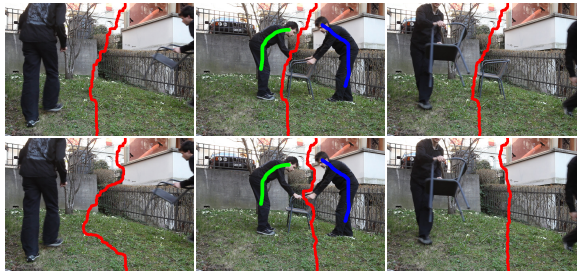

\centering
\graphics{./figures/move4b}\hfill
\graphics{./figures/move5b}\hfill
\graphics{./figures/move6b}\\
\graphics{./figures/move1b}\hfill
\graphics{./figures/move2b}\hfill
\graphics{./figures/move3b}
\caption{Three frames from the \textsc{Chair} sequence. Top, high temporal penalty. Bottom, a low temporal penalty gives more flexibility for the seam to move, allowing it to pass through the chair, which avoids doubling it in the composite.}
\label{fig:movestandard}
\end{figure}

Formally, we compute a labeling that minimizes the following visibility penalty:
\begin{equation}
E = \sum_{i} \left( \sum_{j \in N_s(i)}{\delta(i,j) D(i)} \\
 + \sum_{k \in N_t(i)}{\lambda \delta(i,k) D(i)}\right)
\label{eq:penalty}
\end{equation}
for all pixels $i$, where $D(i) = ||A_i - B_i||^2$ if $A$ and $B$ overlap, 0 otherwise. $N_s(i)$ are the 4 spatial and $N_t(i)$ the 2 temporal neighbors of $i$, and $\delta(i,j)$ is $1$ if $i,j$ are assigned different labels by the seam, else $0$. 
The penalty for cutting temporal neighbors has different significance to cutting spatial neighbors, and is controlled by $\lambda$.
A high $\lambda$ penalizes the seam \emph{moving} over a high cost area, while a low $\lambda$ penalizes the seam cutting through a bad location at each frame. 
We show the effect of changing this parameter in Figure~\ref{fig:movestandard}; with $\lambda$ too high, the seam does not pass through the chair, which is eventually doubled in the output. 
$\lambda=1$ is the default weight used in most examples. 

To minimize $E$, we construct a graph where each node represents a pixel connected in a 3D-grid to spatial and temporal neighbors.
The weights of the edges correspond to the average difference values (\ref{fig:chair3}) of connected nodes.
All pixels with user scribbles are connected to the source or sink respectively, and their weights set to infinity.
Finally, we run a standard min-cut algorithm described by Boykov et al.~\cite{boykov2001experimental} to compute the optimal labeling. 
We now describe two important additions to this basic strategy, motion-compensation of the video volume and a coarse-to-fine refinement, that were necessary for our application.

\begin{figure*}[t]  
    \renewcommand\graphics[1]{\adjincludegraphics[height=6.4em,trim={{.25\width} {0\height} {.15\width} {0\height}},clip]{#1}}
    \newcommand\stackgraphics[1]{\includegraphics[height=7em]{#1}}
    \newcommand\dist{\begin{tikzpicture}[node distance=0.1cm]}
  
  \tikzstyle{line2} = [draw, -latex']
  \centering

  \subfloat[][\label{fig:shift1}regular cube]{\stackgraphics{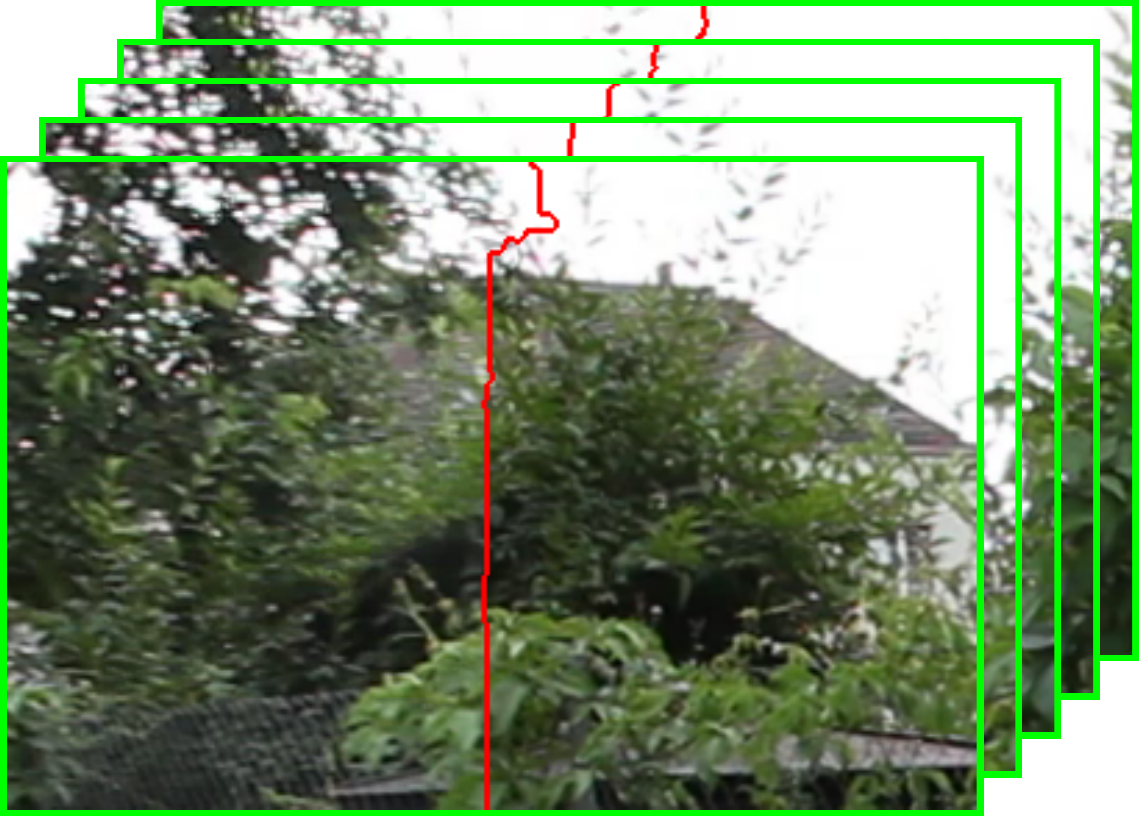}} \ 
  \subfloat[][\label{fig:shift2}motion-compensated cube]{\stackgraphics{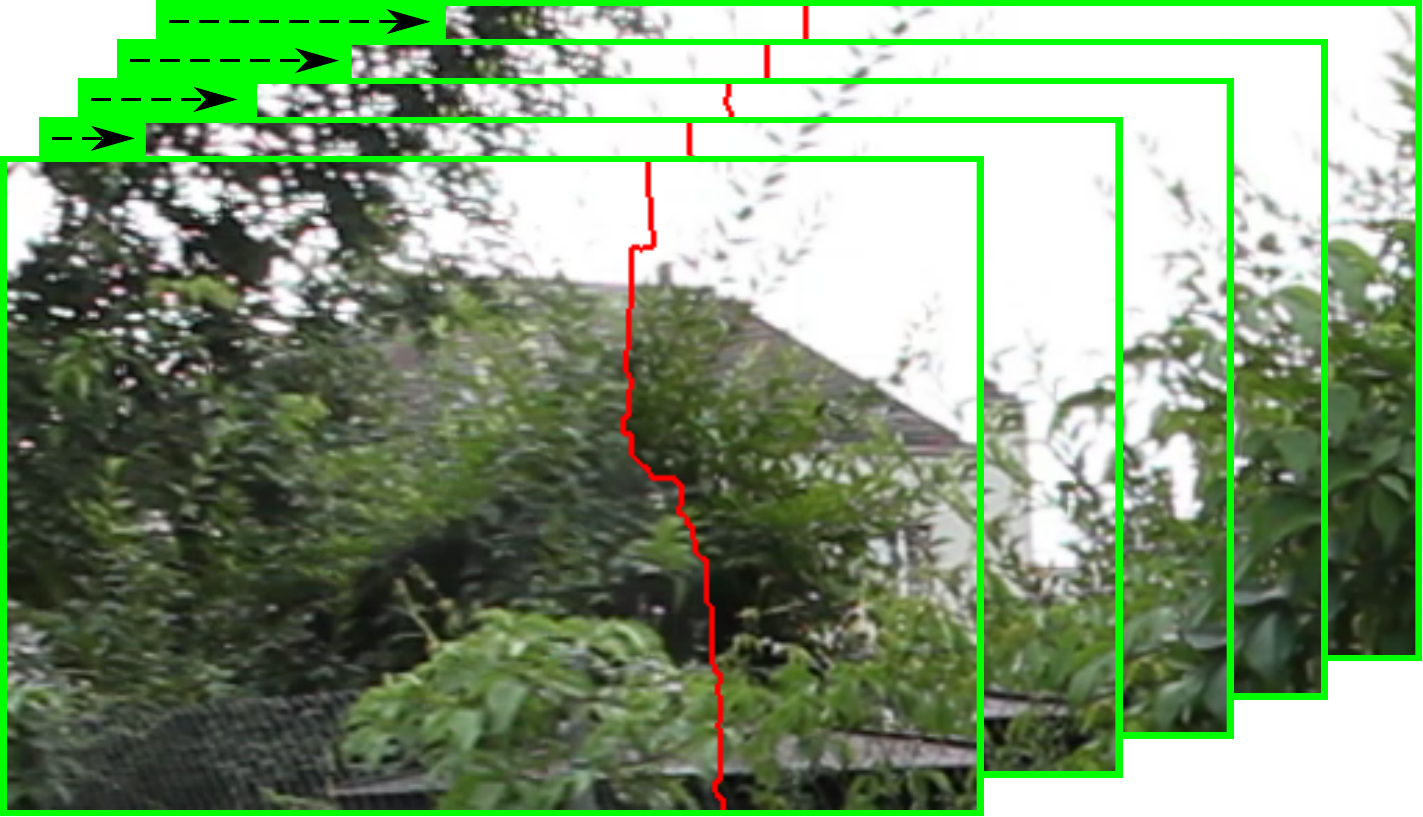}} 
  \subfloat[][\label{fig:shift3}seam moves with frame]{
  \dist
  \node [] (im1) {\graphics{./figures/frame0cut}};
  \node [right=of im1] (im2) {\graphics{./figures/frame40cut}};
  \draw [-latex] (im1.east) |- (im2.west);
  \end{tikzpicture}
  } 
  \subfloat[][\label{fig:shift4}seam moves with content]{
  \dist
  \node [] (im1) {\graphics{./figures/move0cut}};
  \node [right=of im1] (im2) {\graphics{./figures/move40cut}};
  \draw [-latex] (im1.east) |- (im2.west);
  \end{tikzpicture}
  }
  \caption{A seam straight through the video cube stays fixed in the image, but shifts relative to the content~(a,c). Compensating for motion by shifting the video cube (arrows in (b)) causes the seam to move with the content instead~(b,d).}
\end{figure*}

\subsubsect{Motion-Compensated Seams}
The above graph construction assumes that there are no temporal artifacts introduced by a seam that stays in the same location from one frame to the next.
However, this is only true for scenes with a static camera and content.
When the camera or content moves, a stationary seam moves relative to the content (see Figure~\ref{fig:shift3}).
This creates a very disturbing effect, as differences in the background \emph{pop up} when moving over the seam, but are not penalized in the energy formulation above (Equation~\ref{eq:penalty}).

We correct for this by instead constructing the graph such that nodes are connected to their location in the next frame.
One way to find locations in the next frame is by optical flow.
However, optical flow is expensive to compute and prone to errors, and incorrect estimates will cause artifacts in the seam movement.
Instead, we model a \emph{single} global camera motion using homography matrices, which can be robustly estimated (Section~\ref{alignment}).
These are then used to create a \emph{motion-compensated} video cube, where each frame is transformed by the homography mapping it to the previous frame (as seen in Figure~\ref{fig:shift2}).
Then when building the compensated graph, every node in frame $t$ is connected to the node in frame $t+1$ that corresponds to the homography transformation of that point (rounded to the nearest neighbor).

\begin{figure}[t]
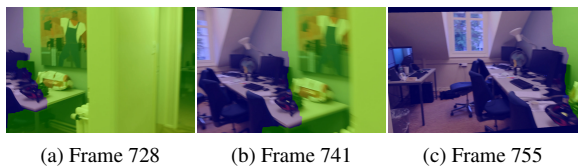

  \renewcommand\mygraphics[1]{\adjincludegraphics[width=.33\linewidth,trim={{.05\width} {0\height} {.1\width} {0\height}},clip]{#1}}

  \centering
  \subfloat[][Frame 728]{\mygraphics{./figures/cutMoving1}}\hfill
  \subfloat[][Frame 741]{\mygraphics{./figures/cutMoving2}}\hfill
  \subfloat[][Frame 755]{\mygraphics{./figures/cutMoving3}}
  \caption{A temporal seam cuts from the first video (green) to the second (blue).  By compensating for camera motion when computing the seam, we can hide the blend in the motion of the camera.}
  \label{fig:cameramove}
\end{figure}

In many cases, this motion-compensation \emph{dramatically} improved the quality of our composites.
Hiding seams in camera motion that are fixed to the content is a perceptually powerful tool that in addition requires only a small region of the content to be well aligned. 
An example can be seen in \textsc{StairsLoop}, and is shown in Figure~\ref{fig:cameramove}, where we blend from one shot to the next while the camera turns into a room.
Since the seam follows the content, it remains undetectable.

\subsubsect{Coarse-to-fine Refinement}
\label{section:coarsetofine}
Doing the above described graph cut on an HD video cube is not computationally feasible.
To solve this problem we reduce the size of the graphs using a coarse-to-fine refinement approach similar to the one described in \cite{lombaert2005multilevel}, but extended to 3D.
In summary, we do the following:

\begin{enumerate}
\item Downsample the video cube (temporally \textit{and} spatially) by a factor of two. Repeated until the resulting cube is small enough for fast processing (3 levels for 1080p video).
\item Run the graph cut on the smallest videocube
\item Grow the region around the seam by $2^{\mathit{grow}}$ pixels. A large \textit{grow} allows more flexibility for cut locations from the previous level. 
\item Upsample this region to the higher resolution cube, and add \emph{only} pixels in the expanded seam region to the new graph.
\item Do the graph cut on the new graph.
\item Repeat from step 3, until the graph cut is run on the full resolution image.
\end{enumerate}

\begin{figure}[t]
\begin{tikzpicture}

\begin{groupplot}[
    group style={
        group name=my fancy plots,
        group size=1 by 2,
        xticklabels at=edge bottom,
        vertical sep=0pt
    },
    width=.99\linewidth,
    xmin=-1, xmax=7
]

\nextgroupplot[ymin=4500,ymax=5200,
             ytick={5000},               
             axis x line=top,
             axis y discontinuity=parallel,
             height=2.8cm]
   
\addplot coordinates {(0, 4954.5)};
\addplot coordinates {(0, 4954.5)};
\addplot coordinates {(0, 4954.5)};
\addplot coordinates {(0, 4954.5)};

\legend{$\mathit{grow}=0$, $\mathit{grow}=1$, $\mathit{grow}=2$, $\mathit{grow}=3$}
\addplot[color=black] coordinates {(0, 4954.5)(0.05, 4700)};
\nextgroupplot[ymin=0,ymax=900,ytickmax=800,               			   
             xtick={0, 1, 2, 3, 4, 5, 6},
             xlabel=\textit{level},
             ylabel=RAM (MiB),               
             axis x line=bottom,
             height=3.5cm]

\addplot coordinates {
(0, 4954.5)
(1, 660.9)
(2, 78.6)
(3, 14.6)
(4, 11.6)
(5, 11)
(6, 10)
};
\addplot coordinates {
(0, 4954.5)
(1,660.9)
(2,78.6)
(3,31.2)
(4,25.2)
(5,25.2)
(6,23.6)
};

\addplot coordinates {
(0, 4954.5)
(1,660.9)
(2,78.6)
(3,60.2)
(4,50.7)
(5,50.7)
(6,50.7)
};

\addplot coordinates {
(0, 4954.5)
(1,660.9)
(2,127.1)
(3,118.7)
(4,98.4)
(5,98.4)
(6,98.4)
};

\end{groupplot}
\end{tikzpicture}
\begin{tikzpicture}
\begin{groupplot}[
    group style={
        group name=my fancy plots,
        group size=1 by 2,
        xticklabels at=edge bottom,
        vertical sep=0pt
    },
    width=.99\linewidth,
    xmin=-1, xmax=7
]

\nextgroupplot[ymin=250,ymax=300,
               ytick={290},
               axis x line=top,
               axis y discontinuity=parallel,
               height=2.8cm]
   
\addplot coordinates {(0, 287.5)};
\addplot coordinates {(0, 287.5)};
\addplot coordinates {(0, 287.5)};
\addplot coordinates {(0, 287.5)};
\addplot[color=black] coordinates {(0, 287.5)(0.01, 260)};

  \legend{$\mathit{grow}=0$, $\mathit{grow}=1$, $\mathit{grow}=2$, $\mathit{grow}=3$}              
  \nextgroupplot[ymin=0,ymax=23,
                 xtick={0, 1, 2, 3, 4, 5, 6},
                 xlabel=\textit{level},
                 axis x line=bottom,
                 ylabel=Time (s),
                 height=3.5cm]

\addplot coordinates {
(0, 287.5)
(1,16.6)
(2,1)
(3,0.5)
(4,0.4)
(5,0.4)
(6,0.5)
};  

\addplot coordinates {
(0, 287.5)
(1,16.6)
(2,1.1)
(3,0.6)
(4,0.5)
(5,0.5)
(6,0.5)
};

\addplot coordinates {
(0, 287.5)
(1,16.9)
(2,1.5)
(3,1)
(4,0.9)
(5,1)
(6,0.9)
};

\addplot coordinates {
(0, 287.5)
(1,18.0)
(2,2.9)
(3,2.2)
(4,1.9)
(5,1.9)
(6,1.9)
};

\end{groupplot}
\end{tikzpicture}
\caption{Time and RAM used for different settings. \textit{level} is the number of coarse-to-fine levels (0 is the standard graph cuts approach), and $2^\mathit{grow}$ is of the search region in pixels added to each side of the seam before upsampling.}
\label{fig:pyramidgraphs}
\end{figure}
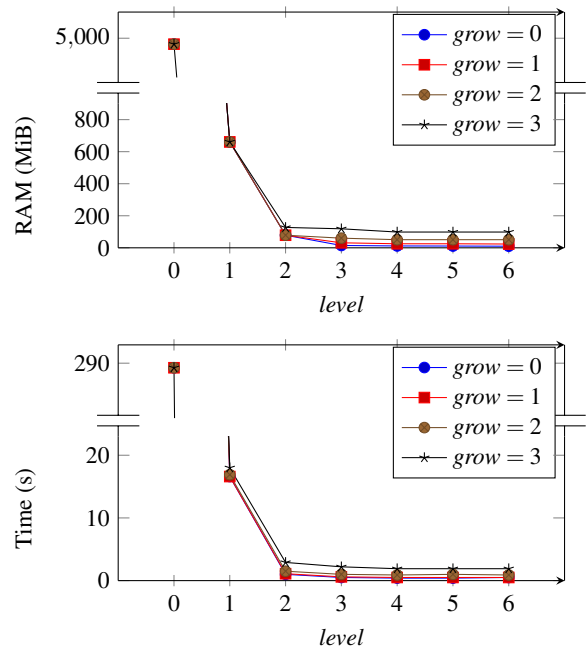

This optimization speeds up the graph cut calculation and reduces memory requirements dramatically.
While actual running times are content dependent, we conducted various runtime and RAM usage experiments on an example clip of ten 1080p frames from the \textsc{Thief} sequence
(Figure~\ref{fig:pyramidgraphs}).
Note the huge speed and memory gains achieved by the coarse-to-fine approach.
On average, using 3 pyramid levels, we compute the video seam at a rate of \textasciitilde 30 ms per frame.

\begin{figure}[tb]
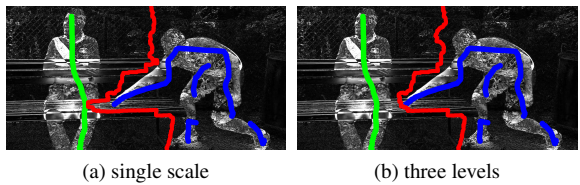

\renewcommand\mygraphics[1]{\adjincludegraphics[width=.49\linewidth,trim={{.2\width} {.1\height} {0\width} {.18\height}},clip]{#1}}
\centering
\subfloat[][single scale]{\mygraphics{./figures/pyramid_difference_0b}}\hfill
\subfloat[][three levels]{\mygraphics{./figures/pyramid_difference_3_4b}}
\caption{Seams (red) computed with different pyramid settings shown over difference image with user strokes (green, blue).}
\label{fig:pyramid_results_short}
\end{figure}

The single scale solution computes a global optimum on the full resolution cube, taking \textasciitilde 5 minutes to compute and using \textasciitilde 5 GiB of RAM.
With the typical settings we used of $\mathit{level}=3$ and $\mathit{grow}=1$ (used for all results shown in this work), we compute the seam in less than a second using only \textasciitilde 30 MiB of RAM. 
Furthermore, that the numerically ``best'' solution computed at the highest resolution can often be semantically poor (see Figure~\ref{fig:pyramid_results_short}); in this case it cuts into the leg of the left person because of similar colors in the foreground and background.
With a coarse-to-fine solution, the seam is required to be good at lower scales of the video volume as well, enforcing consistency of lower frequency content. 

\sect{Alignment}
\label{alignment}
Seam-based compositing works only under the condition that differences between the two takes are small in some regions.
In addition, camera motion must be precisely equalized across takes, or visible wobbling and ``swimming'' artifacts will occur where the content has different motion. 
Even footage filed on a tripod requires alignment, due to small but highly visible single pixel shake.
The solution (and cornerstone of our pipeline) is a robust and fast video alignment technique.

There exist many different solutions for registering/aligning two videos.
After trying the most common approaches, we found limitations in them all, and developed a simple yet robust algorithm that was significantly faster and more stable than other methods we tried.
Many of these out-of-the-box techniques fail due to the large amount of camera shake and significant portions of the image with different content.
\textsc{CatBlurNuke} shows the result of the \textit{F\_Align} node in Nuke, a state-of-the art video compositing tool.
This result took \textasciitilde 7 minutes to compute, compared to our method, which took \textasciitilde 70 seconds, and provides a more robust and stable alignment. 

Alignment can be broken down into two steps; matching, where correspondences are computed between videos, and warping, where one video is warped to the view of the other. 
In this work, we always align the second video to the first. 
Which video is the ``first'' should be chosen by the user, as to maintain artistic intent, we preserve \emph{all} camera motion in this video. 

\subsect{Matching}

\subsubsect{Optical Flow}
While optical flow is designed for temporal frame-to-frame matching, it can also be used to match spatially across views. 
After testing numerous implementations, we found state-of-the-art optical flow methods to be computationally expensive and not able to robustly handle large foreground differences and wide view separation required by our application.

\subsubsect{Features}
Most commonly, images are aligned by matching features (e.g., SURF~\cite{bay2006surf}, or hierarchical Lucas-Kanade~\cite{bouguet2001pyramidal}).
These approaches allow robust outlier rejection and can be efficiently computed, but suffer from missing features in low-texture regions, and dramatically varying features from frame to frame. 
While some approaches compute more regularly spaced features by locally determined feature detection parameters~\cite{grundmann2012calibration}, we observed that even slight differences in feature density from frame to frame can cause strong temporal wobbling after warping (see \textsc{Mirror} example) where RANSAC (Section~\ref{warping}) randomly selects different planes in the world to fit to.
Alternatively, while Structure From Motion (SFM) can find stable 3D points by mapping features to a single 3D model, it cannot handle scenes with low disparity or static/rotating cameras.

\subsubsect{Block-based}
Block-based window-comparison methods consider all pixels evenly and therefore do not have the problem of temporally flickering features.
However, these approaches are most often used for small (temporal) motion, and fail in the case of large scene displacement. 
To compensate for the weaknesses of block-matching and to ensure temporal consistency, we present a coarse-to-fine propagate-and-refine approach to alignment where initial estimates are propagated temporally and then spatially refined. 
We first describe our fast block-matching method, called the hierarchical compass search and then discuss how we use this method to achieve a temporally stable matching. 

\subsubsect{Compass Search}
In its most simple form, we find a single horizontal and vertical offset $dx$, $dy$ that shifts image $B$ to match image $A$:
\begin{equation}
\argmin_{dx,dy}{ \frac{\sum_{x, y}{ \Psi (A(x,y),B(x+dx,y+dy)) }}{\sum_{x, y}{1}}} 
\end{equation}
where $\Psi$ is a distance function; common choices are $L1$ or $L2$ norms, or a combination gradient and color differences in different color spaces (RGB and LAB). 
In all examples here, we use the RGB $L1$ norm, which gave us the best ratio of quality to speed. 
Rather than checking all possible $dx,dy$ (exhaustive search), we use an iterative approach.
Given a $dx,dy$ pair, we test 9 possible neighbor shifts ($dx \pm 1, dy \pm 1$), and then take the one with the minimum difference. 
We then perform a sub-pixel refinement by fitting a parabola through the difference at $dx$ and its two neighbors and computing the extrema.

Of course, this search can only account for motion $\leq$ 1 pixel, and if iterated, it is likely to get stuck at local minima. 
To address both these problems, we find $dx,dy$ using a coarse-to-fine scheme.
We downsample both frames in a pyramid and start the single pixel search on the lowest level, where the pyramid height is determined by the largest expected displacement.
After one step, we double $dx,dy$ and use this as starting point for the next higher resolution level.

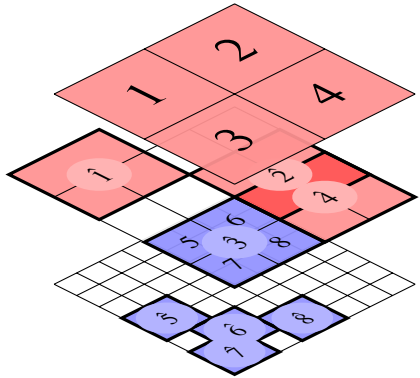
\begin{figure}[tb]
\centering
\begin{tikzpicture}[scale=0.6]
  \tikzstyle{every node}+=[anchor=center,circle]

  \begin{scope}[yslant=0.5,xslant=-1,every node/.append style={yslant=0.5,xslant=-1,font=\footnotesize}]
    \tikzstyle{every node}+=[fill=blue!30]

    \fill[white,fill opacity=0.9] (0,0) rectangle (4,4);

    \fill[blue,fill opacity=0.4] (0,1.5) rectangle (1,2.5);
    \fill[blue,fill opacity=0.4] (0.5,0.5) rectangle (1.5,1.5);
    \fill[blue,fill opacity=0.4] (0,0) rectangle (1,1);
    \fill[blue,fill opacity=0.4] (1.5,0) rectangle (2.5,1);

    \draw[step=0.5, black] (0,0) grid (4,4);
    \draw[black,very thick] (0,1.5) rectangle (1,2.5);
    \draw[black,very thick] (0.5,0.5) rectangle (1.5,1.5);
    \draw[black,very thick] (0,0) rectangle (1,1);
    \draw[black,very thick] (1.5,0) rectangle (2.5,1);

    \node at (0.5, 2) {$\hat5$};
    \node at (1,1) {$\hat6$};
    \node at (0.5, 0.5) {$\hat7$};
    \node at (2,0.5) {$\hat8$};
  \end{scope}

  \begin{scope}[yshift=55, yslant=0.5,xslant=-1,every node/.append style={yslant=0.5,xslant=-1}]
    \fill[white,fill opacity=0.9] (0,0) rectangle (4,4);
    \fill[red,fill opacity=0.4] (0,3) rectangle (2,5);
    \fill[red,fill opacity=0.4] (2,1) rectangle (4,3);
    \fill[blue,fill opacity=0.4] (0,0) rectangle (2,2);
    \fill[red,fill opacity=0.4] (2,0) rectangle (4,2);

    \draw[step=1, black] (0,0) grid (4,4);
    \draw[black,very thick] (0,3) rectangle (2,5);
    \draw[black,very thick] (2,1) rectangle (4,3);
    \draw[black,very thick] (0,0) rectangle (2,2);
    \draw[black,very thick] (2,0) rectangle (4,2);

    \begin{scope}
      \tikzstyle{every node}+=[fill=red!30]
      \node at (1, 4) {$\hat1$};
      \node at (3, 2) {$\hat2$};
      \node at (3, 1) {$\hat4$};
    \end{scope}
    \node[fill=blue!30] at (1, 1) {$\hat3$};

    \node at (1.5, 1.5) {6};
    \node at (1.5, 0.5) {8};
    \node at (0.5, 1.5) {5};
    \node at (0.5, 0.5) {7};
  \end{scope}

  \begin{scope}[yshift=120, yslant=0.5,xslant=-1,every node/.append style={yslant=0.5,xslant=-1,font=\huge}]
    \fill[white,fill opacity=0.9] (0,0) rectangle (4,4);
    \fill[red,fill opacity=0.4] (0,0) rectangle (4,4);

    \draw[step=2, black] (0,0) grid (4,4);

    \node at (1, 3) {1};
    \node at (3, 3) {2};
    \node at (3, 1) {4};
    \node at (1, 1) {3};
  \end{scope}

\end{tikzpicture}
\caption{Hierarchical compass search example. %
Blocks 1-4 search for their best one-pixel displacement ($\hat1$-$\hat4$). 
They are then subdivided at the next level (only block 3 is shown for clarity), and again find optimal displacements ($\hat5$-$\hat8$).}
\label{fig:schematicmatching}
\end{figure}

\begin{figure*}[ht]
  \centering
  
  \renewcommand\mygraphics[1]{\adjincludegraphics[width=.28\linewidth,trim={{0\width} {.1\height} {.2\width} {.1\height}},clip]{#1}}

    \newcommand\myspy[2]{\subfloat[][#1]{
        \begin{tikzpicture}[spy using outlines={magnification=3,size=2.0cm, connect spies, rounded corners}]
          \node { \mygraphics{#2} };
          \spy [blue] on (-1.35,.35) in node [left] at (3, 0);
        \end{tikzpicture}}}

  \myspy{SURF}{./figures/SURF_6_4_1_1} 
  \myspy{OpenCV}{./figures/OpenCV_2_p0005_25000_9_3_200_p005} 
  \myspy{Hierarchical compass search}{./figures/BLOCK_5_5_0} 

  \caption{Example matches (red are outliers rejected by RANSAC). SURF and OpenCV are shown with empirically determined optimal parameters for this dataset.}

  \label{fig:surfcompass}
\end{figure*}
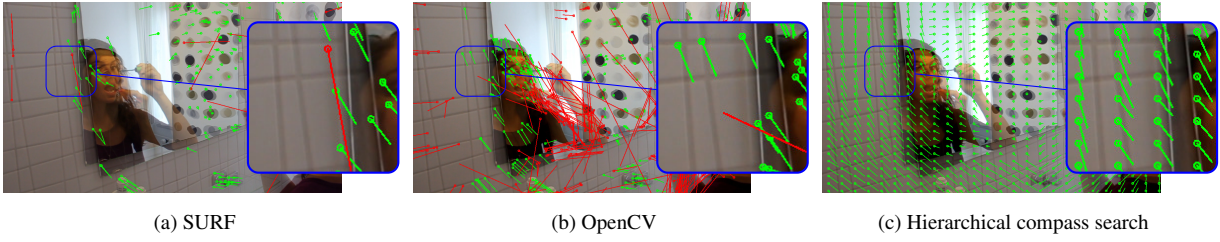

\subsubsect{Hierarchical Compass Search}
The above approach assumes a single, rigid, translation-only model for entire frames, and cannot account for rotations or other homography effects. 
We therefore divide every block into some number of smaller blocks at every level that are then each independently matched.
A diagram of this method can be seen in Figure~\ref{fig:schematicmatching}. 
Displacement vectors from the various approaches are shown in Figure~\ref{fig:surfcompass}. 
Computation on this 720p example took \textasciitilde 66 ms (single-threaded) per frame, considerably less time than SURF (\textasciitilde 1353 ms) and OpenCV's pyramid Lucas-Kanade (\textasciitilde 236 ms), and yielded more stable, reliable results. 

This method has only three, intuitive parameters: $\mathit{smooth}$, $\mathit{level}$, and $\mathit{division}$. 
$\mathit{Level}$ controls the number of coarse-to-fine levels.
$\mathit{Division}$ controls the number of sub-blocks that each block gets divided into at the next level, which determines the final resolution of the matches.
$\mathit{Smooth}$ controls the size of the overlap region in blocks ($smooth=1$ means an overlap of one block). 
We use this hierarchical compass matching to compute both an initial spatial view-to-view matching as well as a per-video temporal frame-to-frame matchings using the default parameters $\mathit{level}=5,\mathit{division}=5,\mathit{smooth}=0$ for nearly all results.
These two matching results form the basis of our temporal propagate-and-refine approach, described in detail in Section~\ref{tempstability}. 

\subsect{Warping}
\label{warping}

While each block only considers translation offsets ($dx,dy$), by aggregating information from multiple blocks, we can estimate more complex warping effects like rotation or perspective change.
We use a homography warp (with RANSAC for outlier rejection) in all our results. 
However, in cases with large amounts of parallax, homographies can be insufficient to model perspective deformation.
We experimented with locally varying image warps to perform alignment~\cite{liu2009content}, a method commonly used in stabilization literature.
However, we found that the high degrees-of-freedom of these warps cause temporal instability that creates very noticeable artifacts when warping between significantly different views (see supplemental example \textsc{FussballWarp}).

\subsect{Temporal Propagation}
\label{tempstability}

Temporally stable alignment is especially important in our application, as we want to both synchronize the global camera motion to that of the first video, and avoid any wobbling from frame to frame.
Our previously described warping can quickly match frames between views, but does not yet give temporally stable results. 
Simply smoothing homography matrices temporally does not work, as we do not know whether the wobbling is due to incorrect estimation or camera shake in the first video (which should be preserved).

Instead, we present a propagate-and-refine solution to address these issues using the hierarchical compass search and warping defined above. 
We first estimate \emph{temporal} homography transformations in the first and second video independently. 
This is much easier than estimating \emph{spatial} alignment, since the image content is very similar, and our hierarchical compass search works without modification. 
We can then exploit the given temporal \emph{and} spatial homographies (visualized as follows): 
\begin{minipage}{\linewidth}
\centering
  \begin{tikzpicture}
    \matrix (m) [matrix of math nodes,row sep=3em,column sep=4em,minimum width=2em]
    {A_{t+1} & B_{t+1} \\
     A_t & B_t \\};

    \draw[-stealth, dashed] (m-1-2) -- node [above] {$H_{t+1}$} (m-1-1);
    \draw[-stealth] (m-2-2) -- node [above] {$H_{t}$} (m-2-1);
    \draw[-stealth] (m-1-1) -- node [left] {$H_a$} (m-2-1);
    \draw[-stealth] (m-1-2) -- node [right] {$H_b$} (m-2-2);
  \end{tikzpicture}
\end{minipage}
\noindent where $A_t$ and $B_t$ are the left and right images at time $t$. 
$H_a,H_b$ are the temporal homography warps, and $H_t$ is the already computed spatial mapping between the two videos (the alignment) for time $t$. 
To get an estimate of $H_{t+1}$ we can simply concatenate the given homographies in the right order: 
\begin{equation}
H_{t+1} = H_a^{-1} * H_t * H_b
\end{equation}

It follows that given $H_a$ and $H_b$ for all frames, we could compute the spatial alignment for a whole video based on one initial alignment ($H_t$) by essentially removing the temporal shake of the second camera, and adding the shake of the first. 
While this indeed reduces wobbling, it introduces drift over time because of inaccuracies in the homography estimation and numerical errors.
Instead, we initialize our hierarchical compass search by transforming the second video with the propagated homography $H_{t+1}$ before matching.
Since the remaining transformation only needs to correct a very small drift, our second-pass block-based approach works very well using the default refinement parameters $\mathit{level}=1,\mathit{division}=4,\mathit{smooth}=1$ in all datasets.

The propagation starts at any frame $t$ where the two images are reasonably close.
This is usually the frame that the user drags to align the videos. 
On that frame $H_t$ is computed, and both forward and backwards $H_a,H_b$ are used to initialize the search for $H_{t+1}$ and $H_{t-1}$ until all frames are aligned. 

This approach forms the basis of our method, and was used in all the results shown.
We allow for an optional iterative refinement step where after the initial cut is computed, the alignment is performed again, but only around the initial cut. 
This is used in cases where the global camera motion does not reflect the motion around the cut, for example when cutting through foreground objects (used only in sequences \textsc{Eyes}, \textsc{Mirror} and \textsc{CinemagraphFussball}).

\sect{Blur Matching}
\label{blur}

\begin{figure}[t]
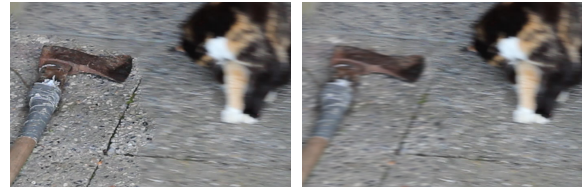

\renewcommand\mygraphics[1]{\adjincludegraphics[width=.49\linewidth,trim={{.22\width} {.22\height} {.42\width} {.35\height}},clip]{#1}}
\centering
\subfloat[][original composite]{
 \label{fig:blur1}
 \mygraphics{./figures/blur1}
}
\subfloat[][after blur matching]{
  \label{fig:blur2}
  \mygraphics{./figures/blur2}
}
\caption{Composition before and after blur matching on sequence \textsc{CatBlur}. In (a), the left half is noticeably sharper.}
\label{fig:blur}
\end{figure}

A common problem of hand-held video footage is camera shake, which introduces motion blur.
With different shake in each take, this blur can occur only in one video at a time, causing the composition to look unnatural after alignment (see Figure~\ref{fig:blur1}).

De-blurring images is a challenging and under-constrained problem where fast and robust solutions are not currently available.
We instead \emph{increase} the blur of the sharper image ($S$) so that it matches the blurrier one ($B$).
There are several common methods to find the local blur in images.
Some calculate a spatially varying Gaussian blur using image pyramids~\cite{trentacoste2011blur} while other methods estimate non-Gaussian blur kernels, for which we refer the reader to a survey paper~\cite{rouf2008study}.

However, all of these techniques are quite slow, and for our application it is sufficient to estimate only a single global blur kernel.
Therefore, we propose a simple and fast blur equalization method. 
We define the ``blurriness'' of an image to be the sum of absolute values of the gradients averaged over color channels, and computed independently for $x$ and $y$ dimensions.
We search the filter kernel space by gradually increasing the size of the blur kernel (starting at 0) until the blurriness of $S$ equals the blurriness of $B$. e.g, for $x$:
\begin{equation}
\argmin_{k_x}{ \sum_{i \in S}{ || \nabla_x F(S,k_x) ||_1 } - \sum_{i \in B}{ || \nabla_x B ||_1 } }
\end{equation}
where $F(S,k_x)$ is the result of filtering image $S$ with kernel of radius $k_x$.
For the choice of filter $F$, we use two varying-width passes of a box filter, which yielded good quality for its computational complexity.
This search is done in two iterations; the first iteration searches different radii $k$ using two passes of a width box filter, and the second iteration updates only the width of the box in the second pass. 
This two-step approach yields reliable convergence and fine scale refinement of kernel shapes, resulting in an optimal triangle-shaped filter. 
Figure~\ref{fig:blur2} shows the improvement after performing blur matching on the composited result.
The automatically computed two-pass box filter kernel size for this image was a 13x3 box filter followed by a 13x1 box filter.
The time to compute the blur kernel depends on the amount of blur; a sequence with a large amount of motion blur (\textsc{BenchMoving}) took \textasciitilde 133 ms per frame to find the kernel, and \textasciitilde 10 ms per frame to apply it.

\sect{Color Matching}
\label{color}

\begin{figure}[tb]
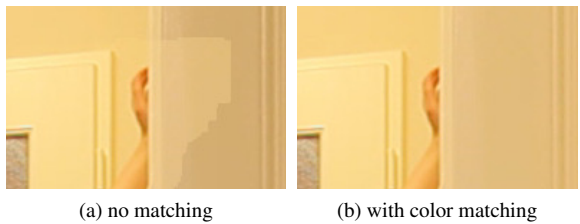


\renewcommand\mygraphics[1]{\adjincludegraphics[width=.49\linewidth,trim={{.65\width} {.72\height} {.05\width} {.05\height}},clip]{#1}}

\renewcommand\mygraphicsb[1]{\adjincludegraphics[width=4cm,trim={{0\width} {.22\height} {0\width} {0\height}},clip]{#1}}

\centering
\subfloat[][no matching]{\label{fig:arms2}\mygraphics{./figures/arms2}}\hfill
\subfloat[][with color matching\label{fig:arms4}]{\mygraphics{./figures/arms4}} \\ \vspace{-.2cm}
\caption{Compositing with (b) and without (a) color matching.}
\label{fig:arms}
\end{figure}

Even when filmed under similar conditions, it is likely that the two takes will differ in brightness, contrast, and hue, due to changes in lighting conditions (see Figure~\ref{fig:arms2}).
This is especially true for cameras that perform automatic white balance and exposure compensation, which often adapt settings while recording.

To address this, we used histogram matching independently on RGB color channels after alignment and before searching for a seam. 
In our application, large differences can exist in foreground content that can bias the histograms. 
We therefore define a similarity threshold ($\gamma$) and build the histograms only out of pixels whose difference is close enough ($||A(x,y)-B(x,y)||_1<\gamma$).
This threshold is quite robust as data is aggregated over the frames, for all results shown we set $\gamma=200$ (for 8-bit images).
It is also very efficient; calculating the lookup tables takes \textasciitilde 12 ms per frame, while the application of the color transform can be done in less than 3 ms.
After the overlap region of the two takes ends (for example with a temporal seam), we slowly fade out the effects of color matching.

\begin{figure}[tb]
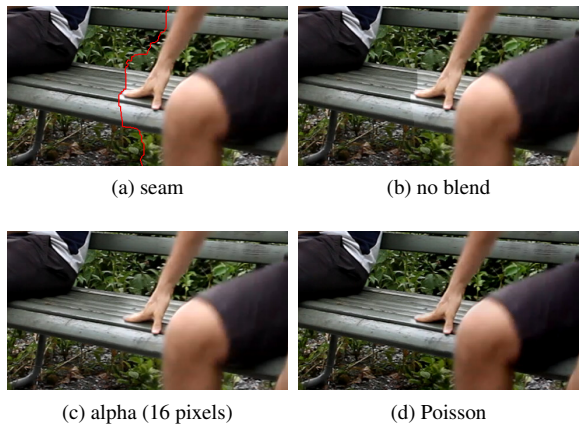


\renewcommand\mygraphics[1]{\adjincludegraphics[width=.49\linewidth,trim={{.35\width} {.3\height} {.4\width} {.45\height}},clip]{#1}}

\centering
\subfloat[][seam]{\mygraphics{./figures/blendnone}}\hfill
\subfloat[][no blend\label{fig:blendnone}]{\mygraphics{./figures/blendnonenocut}}\\
\subfloat[][\label{fig:blendalpha}alpha (16 pixels)]{\mygraphics{./figures/blend4}}\hfill
\subfloat[][\label{fig:blendpoisson}Poisson]{\mygraphics{./figures/blendpoisson}}
\caption{Seam visibility with various blending methods on a difficult example.}
\label{fig:blend}
\end{figure}

\sect{Blending}
\label{blend}
Even after color matching (Section~\ref{color}), the seam can still be visible in a common background. 
This happens when there are non-global illumination differences between the images, like shadows (Figure~\ref{fig:blendnone}).

A fast solution is to perform simple alpha-blending around the seam.
Pixels on the seam are blended equally between the two videos, and the weight falls off linearly in either direction (i.e., $\alpha A + (1-\alpha) B$).
$\alpha$ is computed based on the Manhattan distance to the seam, and the extent of the blend can be user determined (usually between 2 to 32 pixels).
This method can introduce ghosting when the backgrounds are not aligned well enough or foreground objects are close. 
However, since the size of the blending region is usually small, we found that this simple approach worked surprisingly well.
It took on average \textasciitilde 33 ms per frame to compute the Manhattan distance while the actual alpha-blend took \textasciitilde 7 ms per frame. 

For particularly difficult cases, we also added the option of a fast approximation to Poisson blending~\cite{perez2003poisson} using convolution pyramids~\cite{farbman2011convolution}.
While this method improved the result in some cases (see Figure~\ref{fig:blendpoisson}), it created well-known smearing artifacts in many others.
Additionally it is much slower to compute (\textasciitilde 345 ms per frame).
Alpha blending combined with histogram matching was able to remove the effects of lighting, exposure and white balance changes in most scenes, and was used in all results presented.

\sect{Cropping}
\label{crop}

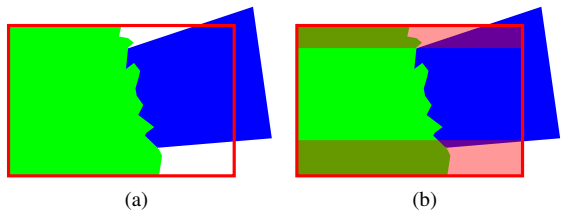
\begin{figure}[tb]
    \newcommand\startfigure{ \begin{tikzpicture}[scale=0.5] }
    \centering
  \subfloat[][\label{fig:cropright}]{
    \startfigure
      \pgfmathsetseed{1234}
      \fill[blue] (1, 0.5) -- (2, 3) -- (6.5, 4.5) -- (7, 1);
      \fill[green] decorate [decoration={random steps,segment length=3pt,amplitude=3pt}]{(4, 0) -- (3, 4)} -- (0, 4) -- (0, 0);
      \draw[red,very thick] (0,0) rectangle (6,4);
    \end{tikzpicture}
  }\hfill \subfloat[][\label{fig:cropborders}]{
    \startfigure
      \pgfmathsetseed{1234}
      \fill[blue] (1, 0.5) -- (2, 3) -- (6.5, 4.5) -- (7, 1);
      \fill[green] decorate [decoration={random steps,segment length=3pt,amplitude=3pt}]{(4, 0) -- (3, 4)} -- (0, 4) -- (0, 0);
      \draw[red,very thick] (0,0) rectangle (6,4);
      \fill[red,even odd rule, opacity=0.4] (0,0) rectangle (6, 4) (0, 0.95) rectangle (6, 3.4);
    \end{tikzpicture}
  }
  \caption{A composite of two videos (green, blue) is contained inside the original video frame (red box) (a).  Our greedy border shrinking approach gives crop~(b).}
\end{figure}

After the alignment, matching, and cut, parts of the final video can be missing (white regions in Figure~\ref{fig:cropright}).
To create an output video, we must crop the final video such that that no missing pixels remain in the whole sequence (if a specific aspect ratio is required, this region can be again cropped).
This is an instance of the ``largest empty cuboid'' problem, where solutions have been shown to have a running time of $O(n^3 + n^2 \log n)$ where $n$ is the number of pixels~\cite{nandy1998maximal} in the volume. 
For our application, missing pixels occur largely at the edges. This observation allows us to present a greedy, iterative solution that runs in $O(ns + nt)$ time where $n$ is the number of pixels in \emph{one} frame, $s$ is bounded by $\max(width,height)$, and $t$ is the number of frames. This is very fast for video sequences, as it propagates the solution across frames, requiring only a minimal number of memory accesses. Our approach does the following:

\begin{figure}[t]
  \mycomment{
  \newcommand\startfigure{ \begin{tikzpicture}[scale=0.58] }
  \centering  
  \subfloat[][\label{fig:cropall}]{
    \startfigure
      \pgfmathsetseed{1234}
      \fill[black, even odd rule] decorate [decoration={random steps,segment length=3pt,amplitude=3pt}]{(5, 0) -- (5.5, 3) -- (5.5, 0)} -- (6, 0) -- (6, 4) -- (0, 4) -- (0, 0) (0.3, 3.2) rectangle (0.5, 3.4);
      \draw[red,very thick] (0,0) rectangle (6,4);

      \fill[red, opacity=0.4] (0,0) -- (3, 2) -- (6, 0);
      \fill[green, opacity=0.4] (0,4) -- (3, 2) -- (6,4);
      \fill[yellow, opacity=0.4] (0,0) -- (3, 2) -- (0,4);
      \fill[blue, opacity=0.4] (6,0) -- (3, 2) -- (6,4);
    \end{tikzpicture}
  }\hfill \subfloat[][\label{fig:cropgood}]{
    \startfigure
      \pgfmathsetseed{1234}
      \fill[black, even odd rule] decorate [decoration={random steps,segment length=3pt,amplitude=3pt}]{(5, 0) -- (5.5, 3) -- (5.5, 0)} -- (6, 0) -- (6, 4) -- (0, 4) -- (0, 0) (0.3, 3.2) rectangle (0.5, 3.4);
      \draw[red,very thick] (0,0) rectangle (6,4);

      \fill[red,even odd rule, opacity=0.4] (0.5,0) rectangle (6, 4) (0, 0) rectangle (5, 4);
    \end{tikzpicture}
  }
  \caption{\label{fig:crophardfig}Each border counts the number of missing pixels in the closest triangle around it (colored)~(a).  Our greedy approach yields the final cropping result~(b).}
  }

\end{figure}

\begin{enumerate}
\item Count the number of empty pixels closest to each border 
\item Choose the border that has the highest number of empty pixels (rightmost border in this example)
\item Crop one pixel from this border
\item Repeat from 1 until no empty pixels are left
\item Initialize the next frame with the current crop
\end{enumerate}

This approach is robust to difficult cases, such as holes and half-islands.
While pathological cases can be constructed where a suboptimal cut is found, we were able to use it successfully for all results presented in this paper.
Furthermore, it is very efficient in practice, requiring only \textasciitilde 2.5 ms per frame.

\chapt{Implementation}

Our user interface has two basic modes.
In the easy mode, almost all parameters are hidden from the end user and are set to their default values. 
This is designed to work out-of-the-box for many tasks, the user must only load two videos, select a temporal alignment by dragging takes relative to each other and start drawing strokes on the frames. 
After that, he or she simply presses the \emph{run} button and the composite is made. 
As needed, more strokes can be added to refine the result. 
For all of the optional blocks (such as blur or color matching), the user decides only whether they should be enabled or not.

The advanced mode exposes all the settings of the individual algorithms.
In this form, one can experiment with other feature matchers and blend settings, tune parameters of the hierarchical compass search, RANSAC, and warping models. 
In general, only cases with particularly difficult camera motion and shake required using the advanced mode; the sequence \textsc{StairsLoop} to tune alignment parameters and the sequences \textsc{Mirror} and \textsc{Eyes} for re-alignment around the cut. 
All the rest of the results (\textasciitilde 90\%) in this paper were generated with the easy mode, meaning that the default values for alignment and matching were used.
All settings, videos, alignment, and cut data can be exported and imported in project files, and our system functions across platforms, allowing for easy collaboration.

\paragraph{Keyframes}
Users can also choose to use keyframes, which divide a single shot into multiple sections where the graph-cut functions independently. 
This allows working on shorter sections of long sequences, which decreases cut computation time, and guarantees that changes between two keyframes will \emph{only} effect that region. 
Keyframes are realized by fixing the seam at each keyframe as a hard constraint in the graph construction. 
This ensures that the seam varies smoothly around keyframes, while still giving independent results in-between them. 
Examples shown here were composed of relatively short takes, and keyframes were not needed. 

\paragraph{Performance}
High performance on FullHD video was one of the key requirements for our system.
Many parts of the program exploit multi-threading, and extensive caching is done to keep the user interface responsive. 
The pipeline seen in Figure~\ref{fig:overview} uses a publisher-subscriber mechanism, where all intermediate data used by the steps above is cached after computation.
When any of the parameters are changed by the user, the system automatically notifies all the dependent nodes and invalidates their cache, recomputing steps as needed in a lazy manner.
This strategy keeps memory requirements low, while reducing the amount of re-computation required.

\chapt{Results and Validation}
\begin{figure}[t]
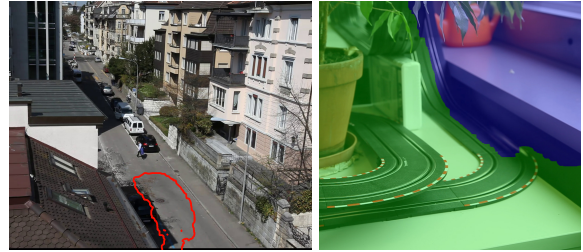

  \renewcommand\mygraphics[1]{\adjincludegraphics[height=3.3cm,trim={{.25\width} {.0\height} {.06\width} {.0\height}},clip]{#1}}
  \centering
  \mygraphics{./figures/result_car} \hfill
  \mygraphics{./figures/result_smallcar}
  \caption{\label{fig:results}Two results from our method (\textsc{EmptyStreet} and \textsc{RaceCar}). Left, a car is removed from the scene by compositing the same location at a different point in time. Right, two clips are combined so that a mistake is hidden.}
\end{figure}

Two seams are shown in Figure~\ref{fig:results}, but as this work is centered around video, we present the majority of our results in supplemental material along with timings and screen casts of their creation.
Here we describe some details in reference to these datasets:

\begin{description}
\item[\textsc{RaceCar} (\textasciitilde 4 min)] 
This example shows DuctTake used for correcting a filming mistake.
In the first take, the car spins off of the racetrack, so a second take is made to continue on from where the mistake happened.

\item[\textsc{Cinemagraph[Fussball / Fountain]} (\textasciitilde 5 / \textasciitilde 3 min)] Our method can also be used out-of-the-box to generate cinemagraphs by simply replacing one video with a static image.

\item[\textsc{PigeonChase} (\textasciitilde 5 min) / \textsc{CatSync} (\textasciitilde 4 min)] Directing animals and children can be very hard.  Using our compositing approach, it is possible to synchronize uncontrollable events, or to remove the trainers.

\item[\textsc{EmptyStreet} (\textasciitilde 6 min)] This sequence shows how a single video can be used in post production to improve a shot (in this case an unwanted car is removed by taking empty street information from the same video at a temporal offset).

\item[\textsc{SpinCombo} (\textasciitilde 8 min)] This video shows a difficult example where a temporal seam must pass through the body of a moving person.
With minor iterations to correct semantic errors of the seam (like doubled balls) we achieve the final result. 

\item[\textsc{BenchMoving} (\textasciitilde 10 min) / \textsc{Church} (\textasciitilde 4 min)] These challenging examples show the robustness of our alignment.
Despite different camera motion, strong perspective effects, and motion blur (in \textsc{BenchMoving}), our motion-compensated graph cut still finds nearly undetectable seams.

\item[\textsc{StairsLoop} (\textasciitilde 12 min) / \textsc{DownSteps} (\textasciitilde 5 min)] These sequences show temporal seams, and are examples of how a motion-compensated cut can be used to hide a temporal transition. \textsc{StairsLoop} contains two cuts, one for the transition into the monitor and the second to loop the video.

\item[\textsc{Mirror} (\textasciitilde 13 min)] This example shows a difficult alignment case where noticeable wobbling occurs due to large camera shake in both takes. 

\end{description}

\paragraph{Comparison to Industry Tools}
We gave four video clips to professional digital artists with the instructions to composite the clips using state-of-the-art industry tools. 
We compare the resulting quality and time required to DuctTake, showing first our time and second the time taken using industry tools.
\begin{description}

\item[\textsc{Chair} (\textasciitilde 2 min / \textasciitilde 25 min)] This sequence is very easy for our system, and the method works without any of the optional blocks seen in Figure~\ref{fig:overview}. Both results (ours and the artist's) look very good. 
Only the shadow of the chair disappears unnaturally in the artist's result.

\item[\textsc{ThroughWindow} (\textasciitilde 3 min / \textasciitilde 30 min)] In the artist's result, a noticeable jump occurs when the composition switches from the first video to the second. Because we have a much more complex temporal seam (visible in the mask) the transition is spread across multiple frames and is less visible.

\item[\textsc{Eyes} (\textasciitilde 4 min / \textasciitilde 10 min)] Both results here are good. Our method requires the additional step of re-aligning around the initial cut.

\item[\textsc{BenchChase} (\textasciitilde 15 min / \textasciitilde 45 min)] In this challenging sequence, the seam has to move very fast near the end.  In the screencast one can see the iterative workflow of our method very well. 

\end{description}
The time spent on all the remaining sequences in the supplementary material varied from 2-7 min per shot.

\chapt{Conclusion and Outlook}

In conclusion, we have presented a workflow and set of algorithms that allow a user to generate high quality composites by computing spatio-temporal seams between videos. 
We described the significant issues that we encountered and presented efficient solutions to these problems.
Our approach is robust and intuitive, and worked over a wide range of examples, most of which were computed with fixed, default parameters. 
However, there are some limitations that prevent certain scenes from working well, and addressing these is an area for future work. 

The most delicate component is alignment; given properly aligned views, we can almost always generate good composites with minimal work. 
Our use of homography warps fails when camera positions are significantly different and the desired seam must pass through objects at different depths. 
See \textsc{Fussball} where wobbling is visible in the narrow left edge of the table, and \textsc{IntoCar} where a jump is visible between clips. 
Additionally, locally varying warping approaches ~\cite{liu2009content} (\textsc{FussballWarp}) added temporal artifacts and visible distortions. 
Fortunately, because our method does not require a complete frame alignment, it is often good enough to have only a small region around the seam be well aligned, for which homography warps can serve well.
In addition, merely positioning cameras carefully circumvents these difficulties and greatly simplifies the alignment process. 

While we focus on compositing two clips, computing a multi-way labeling instead is possible.
However, the energy minimization can then only be approximated by graph cuts (e.g., via alpha expansion~\cite{boykov2001fast}). 
When this effect is desired, we can simply iteratively composite each additional video to the result one at a time (see \textsc{RoofWalk}).

Using seams to composite videos can be fast and robust, but can only be used in restricted cases. 
For example, when desired parts from each take cross over one another, there is no similar region between objects for the seam to cut through. 
These types of scenes would require falling back to the existing approach of rotoscoping objects for segmentation, and pasting them on-top of each other.
As future work, traditional rotoscoping could be incorporated as hard constraints into our graph construction, letting the cut handle the rest of the scene. 

Finally, the numerical minimum found by graph cuts can be different from the conceptual ``best'' seam location, as seams can cut through objects when their color is similar to the background. 
Our method uses iterative strokes to resolve these semantic problems, but interactive seam modification along the lines of panorama weaving~\cite{summa2012panorama} could be a powerful tool.
Unfortunately, the optimizations used in that work to achieve interactive rates do not extend to 3D space-time seams. 
Further speed improvements could be made by utilizing GPU parallelism, as most steps are trivially parallelizable, which could possibly enable this kind of interactive seam modification. 

Despite these limitations we believe that our system introduces a new practical paradigm for video compositing.
As compared to existing state-of-the-art methods, we have shown that this system can produce comparable or better results with much less work.

\section*{Acknowledgments}

We would like to thank Maurizio Nitti for generating the current state-of-the-art results to compare against the newly developed system, and a special thanks to those who appeared in our videos: Iliyan Georgiev, Antoine Milliez, Sarah Pellkofer, Simon Heinzle, Pierre Greisen, and Selina the cat.

\bibliographystyle{eg-alpha}

\bibliography{citations}{}

\end{document}